\documentclass[11pt,a4paper,dvipsnames]{article}
\usepackage[hyperref]{recall_and_learn}
\usepackage{times}
\usepackage{latexsym}

\usepackage{array}
\usepackage{multirow}
\usepackage{amsmath}
\usepackage{amsfonts}
\usepackage{graphicx}
\usepackage{xcolor}
\usepackage{color}
\usepackage{enumerate}
\usepackage{arydshln}
\usepackage{booktabs}
\usepackage{tikz}
\usepackage{algorithm}
\usepackage{subfig}
\usepackage{multirow}
\usepackage[noend]{algorithmic}
\newcommand{\adamcolor}{pink}
\newcommand{\ouradamcolor}{YellowGreen}
\newcommand{\adam}[1]{\colorbox{\adamcolor}{$\displaystyle #1$}}
\newcommand{\adamtext}[1]{\colorbox{\adamcolor}{#1}}
\newcommand{\ouradam}[1]{\colorbox{\ouradamcolor}{$\displaystyle #1$}}
\newcommand{\ouradamtext}[1]{\colorbox{\ouradamcolor}{#1}}
\renewcommand{\algorithmiccomment}[1]{\bgroup\hfill $\triangleright$ ~#1\egroup}

\usepackage{microtype}

\aclfinalcopy 

\title{Recall and Learn: Fine-tuning Deep Pretrained Language Models with Less Forgetting}

\author{
    \parbox{\linewidth}{\centering
	Sanyuan Chen$^{1}$,
	Yutai Hou$^{1}$,
	Yiming Cui$^{1,2}$,
	Wanxiang Che$^{1}$,
	Ting Liu$^{1}$, 
	Xiangzhan Yu$^{1}$
    }%
	\\
	\parbox{\linewidth}{\centering
	$^1$School of Computer Science and Technology, Harbin Institute of Technology, China \\
	$^2$Joint Laboratory of HIT and iFLYTEK Research (HFL), Beijing, China \\
	{\tt \{sychen, ythou, ymcui, car, tliu\}@ir.hit.edu.cn}, 
	{\tt yxz@hit.edu.cn}
	}
}

\date{}

\begin{document}
\maketitle
\begin{abstract}
Deep pretrained language models have achieved great success in the way of pretraining first and then fine-tuning.
But such a sequential transfer learning paradigm often confronts the catastrophic forgetting problem and leads to sub-optimal performance.
To fine-tune with less forgetting, we propose a recall and learn mechanism, which adopts the idea of multi-task learning and jointly learns pretraining tasks and downstream tasks. 
Specifically, we propose a Pretraining Simulation mechanism to recall the knowledge from pretraining tasks without data, and an Objective Shifting mechanism to focus the learning on downstream tasks gradually.
Experiments show that our method achieves state-of-the-art performance on the GLUE benchmark. 
Our method also enables \textit{BERT-base} to achieve better performance than directly fine-tuning of \textit{BERT-large}.
Further, we provide the open-source \textsc{RecAdam} optimizer, which integrates the proposed mechanisms into Adam optimizer, to facility the NLP community.\footnote{https://github.com/Sanyuan-Chen/RecAdam}

\end{abstract}

\section{Introduction}
Deep Pretrained Language Models (LMs), such as ELMo \cite{peters2018deep} and BERT \cite{bert}, have significantly altered the landscape of Natural Language Processing (NLP) and 
a wide range of NLP tasks has been promoted by these pretrained language models. 
These successes are mainly achieved through \textit{Sequential Transfer Learning} \cite{ruder2019neural}:
pretrain a language model on large-scale unlabeled data and then adapt it to downstream tasks. 
The adaptation step is usually conducted in two manners: fine-tuning or freezing pretrained weights.
In practice, fine-tuning is adopted more widely due to its flexibility \cite{smalldata,albert,tuneornot}.

Despite the great success, sequential transfer learning of deep pretrained LMs tends to suffer from catastrophic forgetting during the adaptation step. 
\textit{Catastrophic forgetting} is a common problem for sequential transfer learning, and it happens when a model forgets previously learned knowledge and overfits to target domains \cite{mccloskey1989catastrophic,kirkpatrick2017overcoming}. 
To remedy the catastrophic forgetting in transferring deep pretrained LMs, 
existing efforts mainly explore fine-tuning tricks to forget less. 
ULMFiT \cite{howard2018universal} introduced discriminative fine-tuning, slanted triangular learning rates, and gradual unfreezing for LMs fine-tuning. 
\citet{lee2019mixout} reduced forgetting in BERT fine-tuning by randomly mixing pretrained parameters to a downstream model in a dropout-style.

Instead of learning pretraining tasks and downstream tasks in sequence, \textit{Multi-task Learning} learns both of them simultaneously, thus can inherently avoid the catastrophic forgetting problem.
\citet{xue2019multi} tackled forgetting in automatic speech recognition by jointly training the model with previous and target tasks.  
\citet{kirkpatrick2017overcoming} proposed Elastic Weight Consolidation (EWC) to overcome catastrophic forgetting when continuous learning multiple tasks by adopting the multi-task learning paradigm.
EWC regularizes new task training by constraining the parameters which are important for previous tasks and adapt more aggressively on other parameters. 
Thanks to the appealing effects on catastrophic forgetting, EWC has been widely applied in various domains, such as game playing \cite{ribeiro2019multi}, neural machine translation \cite{thompson2019overcoming} and reading comprehension \cite{xu2019forget}.

However, these multi-task learning methods cannot be directly applied to the sequential transferring regime of deep pretrained LMs. 
Firstly, multi-task learning methods require to use data of pretraining tasks during adaptation, but pretraining data of LMs is often inaccessible or too large for the adaptation. 
Secondly, we only care about the performance of the downstream task, while multi-task learning also aims to promote performance on pretraining tasks.

In this paper, we propose a recall and learn mechanism to cope with the forgetting problem of fine-tuning the deep pretrained LMs. 
To achieve this, we take advantage of multi-task learning by adopting LMs pretraining as an auxiliary learning task during fine-tuning.
Specifically, we propose two mechanisms for the two challenges mentioned above, respectively.
As for the challenge of data obstacles, 
we propose the \textit{Pretraining Simulation} to achieve multi-task learning without accessing to pretraining data.
It helps the model to recall previously learned knowledge by simulating the pretraining objective using only pretrained parameters. 
As for the challenge of learning objective difference, we propose the \textit{Objective Shifting} to balance new task learning and pretrained knowledge recalling.
It allows the model to focus gradually on the new task by shifting the multi-task learning objective to the new task learning. 

We also provide Recall Adam (\textsc{RecAdam}) optimizer to integrate the recall and learn mechanism into Adam optimizer \cite{adam}.
We release the source code of the \textsc{RecAdam} optimizer implemented in Pytorch.
It is easy to use and can facilitate the NLP community for better fine-tuning of deep pretrained LMs.
Experiments on GLUE benchmark with the BERT-base model show that the proposed method can significantly outperform the vanilla fine-tuning method.
Our method with the BERT-base model can even achieve better results than directly fine-tuning the BERT-large model. 
In addition, thanks to the effectiveness of pretrained knowledge recalling, we gain 
better performance by initializing model with random parameters rather than pretrained parameters. 
Finally, we achieve state-of-the-art performance on GLUE benchmark with the ALBERT-xxlarge model.

Our contributions can be summarized as follows: 
(1) We propose to tackle the catastrophic forgetting problem of fine-tuning the deep pretrained LMs by adopting the idea of multi-task learning and obtain state-of-the-art results on GLUE benchmark.
(2) We propose Pretraining Simulation and Objective Shifting mechanisms to achieve multi-task fine-tuning without data of pretraining tasks.
(3) We provide the open-source \textsc{RecAdam} optimizer to facilitate deep pretrained LMs fine-tuning with less forgetting.

\section{Background}
In this section, we introduce two transfer learning settings: sequential transfer learning and multi-task learning.
They both aim to improve the learning performance by transferring knowledge across multiple tasks, but apply to different scenarios.

\subsection{Sequential Transfer Learning}
\label{ssec:stl}

\textit{Sequential transfer learning} learns source tasks and target tasks in sequence,
and transfers knowledge from source tasks to improve the models' performance on target tasks.

It typically consists of two stages: \textit{pretraining} and \textit{adaptation}.
During pretraining, the model is trained on source tasks with the loss function $Loss_{S}$.
During adaptation, the pretrained model is further trained on target tasks with the loss function $Loss_{T}$.
The standard adaptation methods includes \textit{fine-tuning} and \textit{feature extraction}. 
Fine-tuning updates all the parameters of the pretrained model, 
while feature extraction regards the pretrained model as a feature extractor and keeps it fixed during the adaptation phase.

Sequential transfer learning has been widely used recently, and the released deep pretrained LMs have achieved great successes on various NLP tasks \cite{peters2018deep, bert,albert}.
While the adaptation of the deep pretrained LMs is very efficient, it tends to suffer from \textit{catastrophic forgetting}, where the model forgets previously learned knowledge from source tasks when learning new knowledge from target tasks.

\subsection{Multi-task Learning}
\label{ssec:mtl}

\textit{Multi-task Learning} learns multiple tasks simultaneously, and improves the models' performance on all of them by sharing knowledge across these tasks \cite{mtl1997,mtl2017}.

Under the multi-task learning paradigm, the model is trained on both source tasks and target tasks with the loss function: 
\begin{equation}
\label{eq:multitask}
\resizebox{0.7\hsize}{!}{$
        Loss_M = \lambda Loss_{T} + (1-\lambda) Loss_{S} \\
$}
\end{equation}
where $\lambda\in(0, 1) $ is a hyperparameter balancing these two tasks.
It can inherently avoid catastrophic forgetting problem because the loss on source tasks $Loss_{S}$ is always part of the optimization objective.

To overcome catastrophic forgetting problem (discussed in \S~\ref{ssec:stl}), can we apply the idea of multi-task learning to the adaptation of the deep pretrained LMs?
There are two challenges in practice:
\begin{enumerate}[1)]
    \item We cannot get access to the pretraining data to calculate $Loss_{S}$ during adaptation.
    \item The optimization objective of adaptation is $Loss_{T}$, while multi-task learning aims to optimize $Loss_M$, i.e., the weighted sum of $Loss_{T}$ and $Loss_{S}$.
\end{enumerate}

\section{Methodology}
In this section, we introduce Pretraining Simulation (\S~\ref{ssec:pre_simul}) and Objective Shifting (\S~\ref{ssec:obj_shift}) to overcome the two challenges 
(discussed in \S~\ref{ssec:mtl}) respectively.
Pretraining Simulation allows the model to learn source tasks without pretraining data,
and Objective Shifting allows the model to focus on target tasks gradually.
We also introduce the \textsc{RecAdam} optimizer (\S~\ref{ssec:radam}) to integrate these two mechanisms into the common-used Adam optimizer.

\subsection{Pretraining Simulation}
\label{ssec:pre_simul}
As for the first challenge that pretraining data is unavailable, we introduce Pretraining Simulation to approximate the optimization objective of source tasks as a quadratic penalty, which keeps the model parameters close to the pretrained parameters.

Following Elastic Weight Consolidation (EWC; \citealt{kirkpatrick2017overcoming,ewc-deriv}), we approximate the optimization objective of source tasks with Laplace’s Method and independent assumption among the model parameters. Since EWC requires pretraining data, we further introduce a stronger independent assumption and derive a quadratic penalty, which is independent with the pretraining data.
We introduce the detailed derivation process as follows.

From the probabilistic perspective, the learning objective on the source tasks $Loss_{S}$ would be optimizing the negative log posterior probability of the model parameters $\theta$ given data of source tasks $D_S$:
\begin{equation}
\label{eq:post}
\notag
\resizebox{0.47\hsize}{!}{$
    Loss_{S} = - \log p(\theta| D_{S})
$}
\end{equation}

The pretrained parameters $\theta^*$ can be assumed as a local minimum of the parameter space, and it satisfies the equation:
\begin{equation}
\notag
\resizebox{0.6\hsize}{!}{$
    \theta^* = \mathop{\arg\min}_{\theta} \{- \log p(\theta| D_{S}) \}
$}
\end{equation}

Due to the intractability, the optimization objective $- \log p(\theta| D_{S})$ is locally approximated with the Laplace’s Method \cite{laplace}:
\begin{equation}
\label{eq:laplace}
\notag
\resizebox{0.87\hsize}{!}{$
    \begin{aligned}
    - \log p(\theta| D_{S})  \approx 
    &- \log p(\theta^*| D_{S}) \\
    &+ \frac{1}{2} (\theta - \theta^*)^{\top} H(\theta^*) (\theta - \theta^*)
    \end{aligned}
$}
\end{equation}
where $H(\theta^*)$ is the Hessian matrix of the optimization objective w.r.t. $\theta$ and evaluated at $\theta^*$. $- \log p(\theta^*| D_{S})$ is a constant term w.r.t. $\theta$, and it can be ignored during optimization. 

Since the pretrained model convergences on the source tasks, $H(\theta^*)$ can be approximated with the empirical Fisher information matrix $F(\theta^*)$ \cite{efisher}: 
\begin{equation}
\notag
\resizebox{0.95\hsize}{!}{$
    F(\theta^*) = \mathbb{E}_{x \sim D_S} [\nabla_{\theta} \log p_{\theta}(x) \nabla_{\theta} \log p_{\theta}(x)^{\top}|_{\theta = \theta^*}] 
    $} 
\end{equation}
\begin{equation}
\notag
\resizebox{0.6\hsize}{!}{$
    H(\theta^*) \approx N F(\theta^*) + H_{prior}(\theta^*)
$}
\end{equation}
where $N$ is the number of i. i. d. observations in $D_S$, 
$H_{prior}(\theta^*)$ is the Hessian matrix of the negative log prior probability
$- \log p(\theta)$.

Because of the computational intractability, 
EWC approximate $H(\theta^*)$ by using the diagonal of $F(\theta^*)$ and ignoring the prior Hessian matrix $H_{prior}(\theta^*)$:
\begin{equation}
\label{eq:ewc}
\notag
\resizebox{0.85\hsize}{!}{$
    (\theta - \theta^*)^{\top} H(\theta^*) (\theta - \theta^*) 
    \approx  N \sum_{i} F_i (\theta_i - \theta_{i}^*)^2 
$}
\end{equation}
where $F_i$ is the corresponding diagonal Fisher information value of the model parameter $\theta_i$.

Since the pretraining data is unavailable, we further approximate $H(\theta^*)$ with a stronger assumption that each diagonal Fisher information value $F_i$ is independent of the corresponding parameter $\theta_i$:
\begin{equation}
\label{eq:our}
\notag
\resizebox{0.85\hsize}{!}{$
    (\theta - \theta^*)^{\top} H(\theta^*) (\theta - \theta^*) 
    \approx  N F \sum_{i} (\theta_i - \theta_{i}^*)^2 
$}
\end{equation}

The final approximated optimization objective of the source tasks is the quadratic penalty between the model parameters and the pretrained parameters:
\begin{equation}
\label{eq:final}
\notag
\resizebox{0.93\hsize}{!}{$
    \begin{aligned}
    Loss_S 
    &=- \log p(\theta| D_{S})  \\
    & \approx \frac{1}{2} (\theta - \theta^*)^{\top} H(\theta^*) (\theta - \theta^*) \\
    &\approx \frac{1}{2} (\theta - \theta^*)^{\top} (N F(\theta^*) + H_{prior}(\theta^*)) (\theta - \theta^*) \\
    &\approx \frac{1}{2} N \sum_{i} F_i (\theta_i - \theta_{i}^*)^2 \\
    &\approx \frac{1}{2} N F \sum_{i} (\theta_i - \theta_{i}^*)^2 \\
    &= \frac{1}{2} \gamma \sum_{i} (\theta_i - \theta_{i}^*)^2
    \end{aligned}
$}
\end{equation}
where $ \frac{1}{2} \gamma$ is the coefficient of the quadratic penalty.

\subsection{Objective Shifting}
\label{ssec:obj_shift}
As for the second challenge that the optimization objective of multi-task learning is inconsistent with adaptation,
we introduce Objective Shifting to allow the objective function to gradually shift to $Loss_T$ with the annealing coefficient.

We replace the coefficient $\lambda$ in the optimization objective of multi-task learning (as shown in Eq.~\ref{eq:multitask}) with the annealing function $\lambda(t)$, where $t$ refers to the update timesteps during fine-tuning. The loss function of our method is set to multi-task learning with annealing coefficient:
\begin{equation}
\notag
\resizebox{0.8\hsize}{!}{$
    Loss = \lambda(t) Loss_{T} + (1-\lambda(t)) Loss_{S}
$}
\end{equation}

\begin{figure}[t]
	\centering
	\begin{tikzpicture}
	\draw (0,0 ) node[inner sep=0] {\includegraphics[height=6.cm]{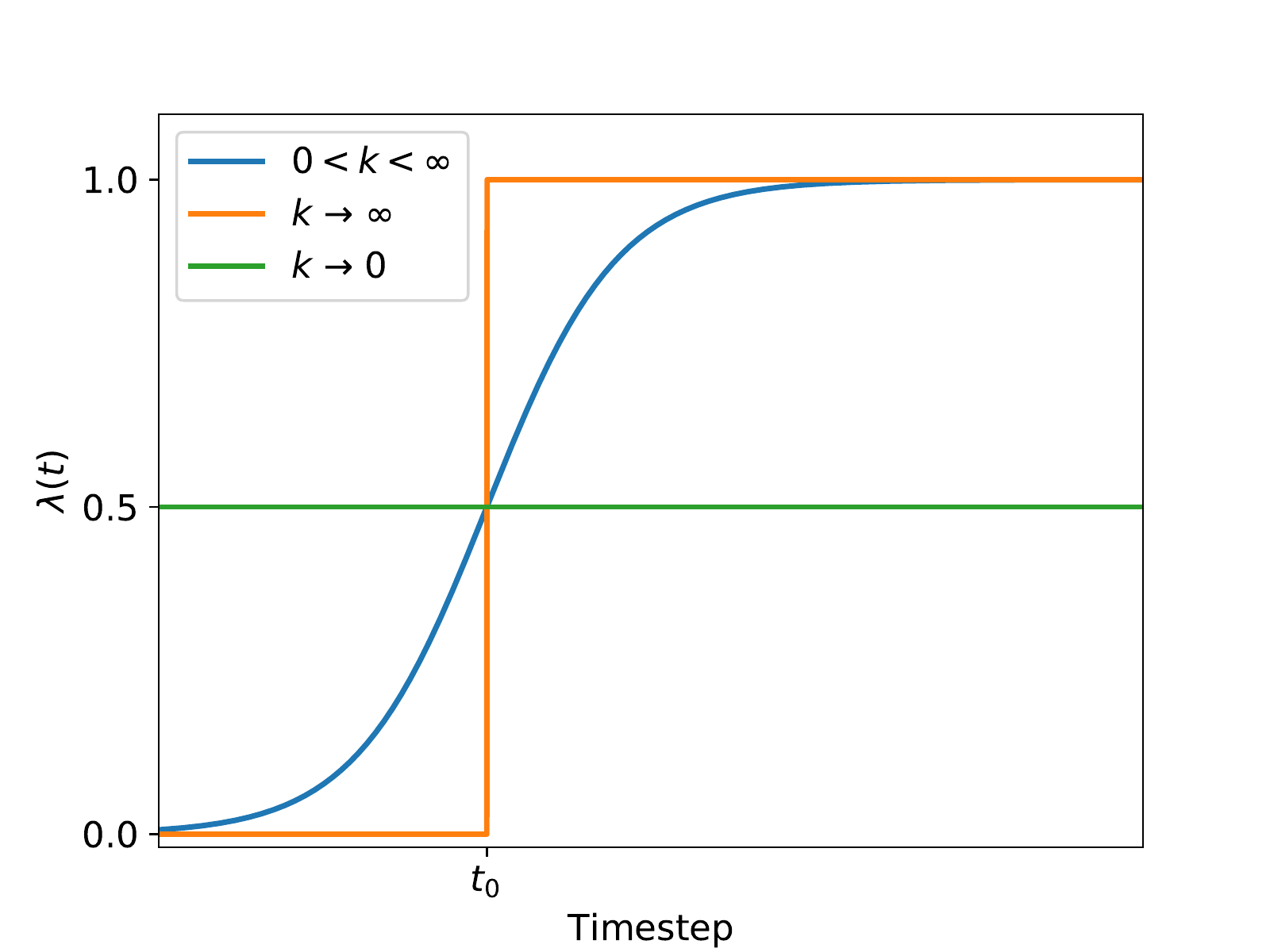}};
	\end{tikzpicture}
	\caption{
	Objective Shifting: we replace the coefficient $\lambda$ with the annealing function $\lambda(t)$. Fine-tuning and multi-task learning can be regarded as the special cases ($k \rightarrow \infty$ and $k \rightarrow 0$) of our method.
		 }\label{fig:anneal}
\end{figure}

Specifically, to better balance the multi-task learning and fine-tuning, $\lambda(t)$ is calculated as the sigmoid annealing function \cite{svae}:

\begin{equation}
\notag
    \lambda(t) = \frac{1}{1 + \exp (- k \cdot (t - t_0))}
\end{equation}
where $k$ and $t_0$ are the hyperparameters controlling the annealing rate and timesteps. 

As shown in Figure~\ref{fig:anneal}, at the beginning of the training process, 
the model mainly learns general knowledge by focusing more on pretraining tasks.
As training progress, the model gradually focuses on target tasks and learns more target-specific knowledge while recalling the knowledge of pretraining tasks. 
At the end of the training process, the model completely focuses on target tasks, and the final optimization objective is $Loss_{T}$.

Fine-tuning and multi-task learning can be regarded as special cases of our method.
When $k \rightarrow \infty$, our method can be regarded as fine-tuning. The model firstly gets pretrained on source tasks with the $Loss_{S}$, then learns the target tasks with the $Loss_{T}$. 
When $k \rightarrow 0$, $\lambda(t)$ is a constant function, then our method can be regarded as the multi-task learning. The model learns source tasks and target tasks simultaneously with the loss function $ \frac{1}{2} (Loss_{T} + Loss_{S})$.

\begin{algorithm*}[tb!]
\caption{\adamtext{Adam} and \ouradamtext{RecAdam}}
\footnotesize
\label{algo_adamr}
\begin{algorithmic}[1]
\STATE{\textbf{given} initial learning rate $\alpha \in \mathbb{R}$,  momentum factors $\beta_1 = 0.9, \beta_2 =0.999, \epsilon = 10^{-8}$, pretrained parameter vector $\theta^* \in \mathbb{R}^n$, coefficient of quadratic penalty $\gamma \in \mathbb{R}$, annealing coefficient in objective function $\lambda(t) = 1 / (1 + \exp(-k \cdot (t - t_0)), k \in \mathbb{R}, t_0 \in \mathbb{N}$} \label{adam-Given} 
\STATE{\textbf{initialize} timestep $t \leftarrow 0$, parameter vector $\theta_{t=0} \in \mathbb{R}^n$,  first moment vector $m_{t=0} \leftarrow 0$, second moment vector  $v_{t=0} \leftarrow 0$, schedule multiplier $\eta_{t=0} \in \mathbb{R}$}
\REPEAT
	\STATE{$t \leftarrow t + 1$}
	\STATE{$\nabla f_t(\theta_{t-1}) \leftarrow  \text{SelectBatch}(\theta_{t-1})$}  \COMMENT{select batch and return the corresponding gradient}
	\STATE{$g_t \leftarrow \adam{\lambda(t)} \nabla f_t(\theta_{t-1})$  \adam{+ (1-\lambda(t)) \gamma (\theta_{t-1}-\theta^*)}}\label{adam_imp}
	\STATE{$m_t \leftarrow \beta_1 m_{t-1} + (1 - \beta_1) g_t $} \label{adam-mom1} \COMMENT{here and below all operations are element-wise}
	\STATE{$v_t \leftarrow \beta_2 v_{t-1} + (1 - \beta_2) g^2_t $} \label{adam-mom2}
	\STATE{$\hat{m}_t \leftarrow m_t/(1 - \beta_1^t) $} \COMMENT{$\beta_1$ is taken to the power of $t$} \label{adam-corr1}
	\STATE{$\hat{v}_t \leftarrow v_t/(1 - \beta_2^t) $} \COMMENT{$\beta_2$ is taken to the power of $t$} \label{adam-corr2}
	\STATE{$\eta_t \leftarrow \text{SetScheduleMultiplier}(t)$}	\COMMENT{can be fixed, decay, or also be used for warm restarts}\label{setschedule}
	\STATE{$\theta_t \leftarrow \theta_{t-1} - \eta_t \left( \ouradam{\lambda(t)} \alpha  \hat{m}_t / (\sqrt{\hat{v}_t} + \epsilon) \ouradam{+ (1-\lambda(t))  \gamma (\theta_{t-1}-\theta^*)} \right)$} \label{adam-xupdate}
\UNTIL{ \textit{stopping criterion is met} }
\RETURN{optimized parameters $\theta_t$}
\end{algorithmic}
\end{algorithm*}

\subsection{RecAdam Optimizer}
\label{ssec:radam}
Adam optimizer \cite{adam} is commonly used for fine-tuning the deep pretrained LMs. 
We introduce Recall Adam (\textsc{RecAdam}) optimizer to integrate the quadratic penalty and the annealing coefficient, which are the core factors of the Pretraining Simulation (\S~\ref{ssec:pre_simul}) and Objective Shifting (\S~\ref{ssec:obj_shift}) mechanisms respectively, by decoupling them from the gradient updates in Adam optimizer.

\citet{adamw} observed that L2 regularization and weight decay are not identical for adaptive gradient algorithms such as Adam, and confirmed the proposed AdamW optimizer based on decoupled weight decay could substantially improve Adam's performance in both theoretical and empirical way. 

Similarly, it is necessary to decouple the quadratic penalty and the annealing coefficient when fine-tuning the pretrained LMs with Adam optimizer.
Otherwise, both the quadratic penalty and annealing coefficient would be adapted by the gradient update rules, resulting in different magnitudes of the quadratic penalty among the model's weights.

The comparison between Adam and \textsc{RecAdam} are shown in Algorithm~\ref{algo_adamr}, where SetScheduleMultiplier$(t)$ (Line~\ref{setschedule}) refers to the procedure (e.g. warm-up technique) to get the scaling factor of the step size.

Line~\ref{adam_imp} of Algorithm~\ref{algo_adamr} shows how we implement the quadratic penalty and annealing coefficient with the vanilla Adam optimizer. The weighted sum of the gradient of target task objective function $\nabla f(\theta)$ and the gradient of the quadratic penalty $\gamma (\theta-\theta^*)$ get adapted by the gradient update rules, which derives to inequivalent magnitudes of the quadratic penalty among the model's weights, e.g. the weights that tend to have larger gradients $\nabla f(\theta)$ would have the larger second moment $v$ and be penalized by the relatively smaller amount than other weights.

With \textsc{RecAdam} optimizer, we decouple the gradient of the quadratic penalty $\gamma (\theta-\theta^*)$ and the annealing coefficient $\lambda(t)$ in Line~\ref{adam-xupdate} of Algorithm~\ref{algo_adamr}. In this way, only the gradient of target task objective function $\nabla f(\theta)$ get adapted during the optimization steps, and all the weights of the training model would be more effectively penalized with the same rate $(1-\lambda(t))  \gamma$. 

Since the \textsc{RecAdam} optimizer is only one line modification from Adam optimizer, it can be easily used by feeding the additional parameters, including the pretrained parameters and a few hyperparameters of the Pretraining Simulation and Objective Shifting mechanisms.

\section{Experiments}

\subsection{Setup}

\paragraph{Model}
We conduct the experiments with the deep pretrained language model BERT-base \cite{bert} and ALBERT-xxlarge \cite{albert}.

BERT is a deep bi-directional pretrained model based on multi-layer Transformer encoders. It is pretrained on the large-scale corpus with two unsupervised tasks:  Masked LM and Next Sentence Prediction, and has achieved significant improvements on a wide range of NLP tasks. We use the BERT-base model with 12 layers, 12 attention heads and 768 hidden dimensions.

ALBERT is the latest deep pretrained LM that achieves the state-of-the-art performance on several benchmarks. It improves BERT by the parameter reduction techniques and self-supervised loss for sentence-order prediction (SOP). The ALBERT-xxlarge model with 12 layers, 64 attention heads, 128 embedding dimension and 4,096 hidden dimensions is the current state-of-the-art model released by \citet{albert}.

\paragraph{Data}
We evaluate our methods on the General Language Understanding Evaluation (GLUE) benchmark \cite{glue}.

GLUE is a well-known benchmark focused on evaluating model capabilities for natural language understanding. It includes 9 tasks: Corpus of Linguistic Acceptability (CoLA; \citealt{cola}), Stanford Sentiment Treebank (SST; \citealt{sst}), Microsoft Research Paraphrase Corpus (MRPC; \citealt{mrpc}), Semantic Textual Similarity Benchmark (STS; \citealt{sts}), Quora Question Pairs (QQP; \citealt{qqp}), Multi-Genre NLI (MNLI; \citealt{mnli}), Question NLI (QNLI; \citealt{qnli}), Recognizing Textual Entailment (RTE; \citealt{rte-1,rte-2,rte-3,rte}) and Winograd NLI (WNLI; \citealt{wnli}).

Following previous works \cite{xlnet,roberta,albert}, we report our single-task single-model results on the dev set of 8 GLUE tasks, excluding the problematic WNLI dataset.\footnote{https://gluebenchmark.com/faq} We report Pearson correlations for STS, Matthew's correlations for CoLA, the “match” condition (MNLI-m) for MNLI, and accuracy scores for other tasks. 

\paragraph{Implementation}
As discussed in \S~\ref{ssec:radam}, we implement the Pretraining Simulation and Objective Shifting techniques with the proposed \textsc{RecAdam} optimizer. 
Our methods use random initialization because of the pretrained knowledge recalling implementation, while vanilla fine-tuning initializes the fine-tuning model with pretrained parameters.

We fine-tune BERT-base and ALBERT-xxlarge model with the same hyperparameters following \citet{bert} and \citet{albert}, except for the maximum sequence length which we set to 128 rather than 512. 
For the BERT-base model, we set the learning rate to 2e-5 and select the training step to make sure the convergence of vanilla fine-tuning on each target task. 
We note that we fine-tune for RTE, STS, and MRPC directly using the pretrained LM while the previous works are using an MNLI checkpoint for further performance improvement.
As for the hyperparameters of our methods, we set $\gamma$ in the quadratic penalty to 5,000, and select the best $t_0$ and $k$ in \{100, 250, 500, 1,000\} and \{0.05, 0.1, 0.2, 0.5, 1\} respectively for the annealing coefficient $\lambda(t)$. 
Following previous works \cite{xlnet,roberta,albert}, we report the score of 5 differently-seeded runs for each result.

\begin{table*}[t]
	\centering
	\footnotesize
	\setlength{\tabcolsep}{1.4mm}
	\begin{tabular}{lccccccccccc}
		\toprule
		\multirow{2}{*}{\textbf{Model}} &
		\textbf{MNLI} & \textbf{QQP} & \textbf{QNLI} & \textbf{SST} & \textbf{Avg} & \textbf{CoLA} & \textbf{STS} & \textbf{MRPC} & \textbf{RTE} & \textbf{Avg} & \multirow{2}{*}{\textbf{Avg}} \\ &
		392k & 363k & 108k & 67k & $>$10k & 8.5k & 5.7k & 3.5k & 2.5k & $<$10k  \\ 
		\midrule
		BERT-base \cite{bert} & 84.4 & - & 88.4 &  92.7 & - & - & - & 86.7 & - & - & -   \\ 
		BERT-base (rerun) $_{Median}$ & 84.8 & \textbf{91.4} & 88.6 & 93.0 & 89.5 & 60.6 & 89.8 & 86.5 & 71.1 & 77.0 & 83.2   \\
        BERT-base + RecAdam $_{Median}$ & \textbf{85.3} & \textbf{91.4} & \textbf{89.1} & \textbf{93.6} & \textbf{89.9} 
        & \textbf{62.4} & \textbf{90.4} & \textbf{87.7} & \textbf{74.4} & \textbf{78.7} & \textbf{84.3} \\
        
        \midrule
        BERT-base (rerun) $_{Max}$ & 85.2 & 91.4 & 89.0 & 93.3 & 89.7 & 61.6 & 89.9 & \textbf{88.7} & 71.5 & 77.9 & 83.8   \\
        BERT-base + RecAdam $_{Max}$ & \textbf{85.4} & \textbf{91.6} & \textbf{89.4} & \textbf{94.0} & \textbf{90.1} 
        & \textbf{62.6} & \textbf{90.6} & \textbf{88.7} & \textbf{77.3} & \textbf{79.8} & \textbf{85.0} \\
        
		\midrule
		BERT-large \cite{bert} & 86.6 & 91.3 & 92.3 & 93.2 & 90.9 & 60.6 & 90.0 & 88.0 & 70.4 & 77.3 & 84.1 \\ 
		XLNet-large \cite{xlnet} & 89.8 & 91.8 & 93.9 & 95.6 & 92.8 & 63.6 & 91.8 & 89.2 & 83.8 & 82.1 & 87.4\\
		RoBERTa-large \cite{roberta} & 90.2 & 92.2 & 94.7 & 96.4 & 93.4 & 68.0 &  92.4 & 90.9 & 86.6 & 84.5 & 88.9 \\
		ALBERT-xxlarge \cite{albert} & \textbf{90.8} & 92.2 & 95.3 & \textbf{96.9} & \textbf{93.8} & 71.4 & \textbf{93.0} & 90.9 & 89.2 & 86.1 & 90.0\\
		ALBERT-xxlarge (rerun) $_{Median}$ & 90.6 & 92.2 & \textbf{95.4} & 96.7 & 93.7
		& 69.5 & \textbf{93.0} & 91.2 & 87.4 & 85.3 & 89.5\\ 
        ALBERT-xxlarge + RecAdam $_{Median}$ &  90.5 & \textbf{92.3} & 95.3 & 96.8 & 93.7
        & \textbf{72.9} & 92.9 & \textbf{91.9} & \textbf{89.3} & \textbf{86.8} & \textbf{90.2} \\ 
        
        \midrule
        ALBERT-xxlarge (rerun) $_{Max}$ & \textbf{90.7} & 92.2 & 95.4 & 96.8 & 93.8
		& 72.1 & \textbf{93.2} & 91.4 & 89.9 & 86.7 & 90.2\\ 
        ALBERT-xxlarge + RecAdam $_{Max}$ &  90.6 & \textbf{92.4} & \textbf{95.5} & \textbf{97.0} & \textbf{93.9}
        & \textbf{75.1} & 93.0 & \textbf{93.1} & \textbf{91.7} & \textbf{88.2} & \textbf{91.1} \\ 

    	\bottomrule
	\end{tabular}
	\caption{State-of-the-art single-task single-model results on the dev set of the GLUE benchmark. 
	The number below each task refers to the number of training data. 
	The average scores of the tasks with large training data ($>$10k), the tasks with small training data ($<$10k), and all the tasks are reported separately.
    We rerun the baseline of vanilla fine-tuning without further pretraining on MNLI.
	We report median and maximum over 5 runs.
	}\label{tbl:sota}
\end{table*}

\subsection{Results on GLUE}
Table~\ref{tbl:sota} shows the single-task single-model results of our \textsc{RecAdam} fine-tuning method comparing to the vanilla fine-tuning method with BERT-base and ALBERT-xxlarge model on the dev set of the GLUE benchmark.

\paragraph{Results with BERT-base}
With the BERT-base model, we outperform the vanilla fine-tuning method on 7 out of 8 tasks of the GLUE benchmark and achieve 1.1\% improvements on the average median performance.

Especially for the tasks with smaller training data ($<$10k), our method can achieve significant improvements (+1.7\% on average) compared to the vanilla fine-tuning method.
Because of the data scarcity, vanilla fine-tuning on these tasks are potentially brittle, and rely on the pretrained parameters to be reasonably close to an ideal setting for the target task \cite{smalldata}. 
With the proposed \textsc{RecAdam} method, we successfully achieve better fine-tuning by learning target tasks while recalling the knowledge of pretraining tasks.

It is interesting to find that compared to the median results with BERT-large model, we can also achieve better results on more than half of the tasks (e.g., +4.0\% on RTE, +0.4\% on STS, +1.8\% on CoLA, +0.4\% on SST, +0.1\% on QQP) and better average results (+0.2\%) of all the GLUE tasks. Thanks to the less catastrophic forgetting realized by \textsc{RecAdam}, we can get comparable overall performance with much fewer parameters of the pretrained model.

\paragraph{Results with ALBERT-xxlarge}
With the state-of-the-art model ALBERT-xxlarge, we outperform the vanilla fine-tuning method on 5 out of 8 tasks of the GLUE benchmark and achieve the state-of-the-art single-task single-model average median performance 90.2\% on dev set of the GLUE benchmark. 

Similar to the results with the BERT-base model, We find that our improvements mostly come from the tasks with smaller training data ($<$10k), and we can improve the ALBERT-xxlarge model's median performance on these tasks by +1.5\% on average. Also, compared to the reported results by \citet{albert}, we can achieve similar or better median results on RTE (+0.1\%), STS (-0.1\%), and MRPC (+1.0\%) tasks without pretraining on MNLI task. 

Overall, we outperform the average median results of the baseline with the ALBERT-xxlarge model by 0.7\%, which is lower than the improvement we gain with the BERT-base model (+1.1\%). With advanced model design and pretraining techniques, ALBERT-xxlarge achieves significantly better performance on GLUE benchmark, which would be harder to be further improved. 

\subsection{Analysis}

\begin{table}[t]
	\centering
	\footnotesize
	\setlength{\tabcolsep}{1.5mm} 
	\begin{tabular}{lcccccc}
		\toprule
		{\textbf{Method}} &
		{\textbf{CoLA}} & {\textbf{STS}} & {\textbf{MRPC}} & {\textbf{RTE}} & \textbf{Avg} \\
		\midrule
		vanilla fine-tuning & 60.6 & 89.8 & 86.5 & 71.1 & 77.0 \\
		RecAdam + PI & 62.0 & \textbf{90.4} & 87.3 & 73.6 & 78.3 \\
		RecAdam + RI & \textbf{62.4} & \textbf{90.4} & \textbf{87.7} & \textbf{74.4} & \textbf{78.7} \\
    	\bottomrule
	\end{tabular}
	\caption{Comparison of different model initialization strategies: pretrained initialization (PI) and Random Initialization (RI). We report median over 5 runs.
	}\label{tbl:init}
\end{table}

\paragraph{Model Initialization}
With our \textsc{RecAdam} method based on Pretraining Simulation and Objective Shifting, the model can be initialized with random values, and recall the knowledge of pretraining tasks while learning the new tasks.

It is interesting to see whether the choice of initialization strategies would have an impact on the performance of our \textsc{RecAdam} method. 
Table~\ref{tbl:init} shows the performance comparison of different initialization strategies for \textsc{RecAdam} obtained by the BERT-base model.
It shows that \textsc{RecAdam}, with both initialization strategies, can outperform the vanilla fine-tuning method on all the four tasks.
For the target task STS,
the model with pretrained initialization can achieve the same result as random initialization.
For the other tasks (e.g., CoLA, MRPC, RTE), Random initialize the model would be our best choice. It is because the model would benefit from a larger parameter search space with random initialization. In contrast, with pretrained initialization, the search space would be limited to around the pretraining model, making it harder for the model to learn the new tasks.

\begin{figure*}[tbp]
	\centering
	\subfloat[Training loss on the target task]
	{\label{sfig:a}\includegraphics[scale=0.4]{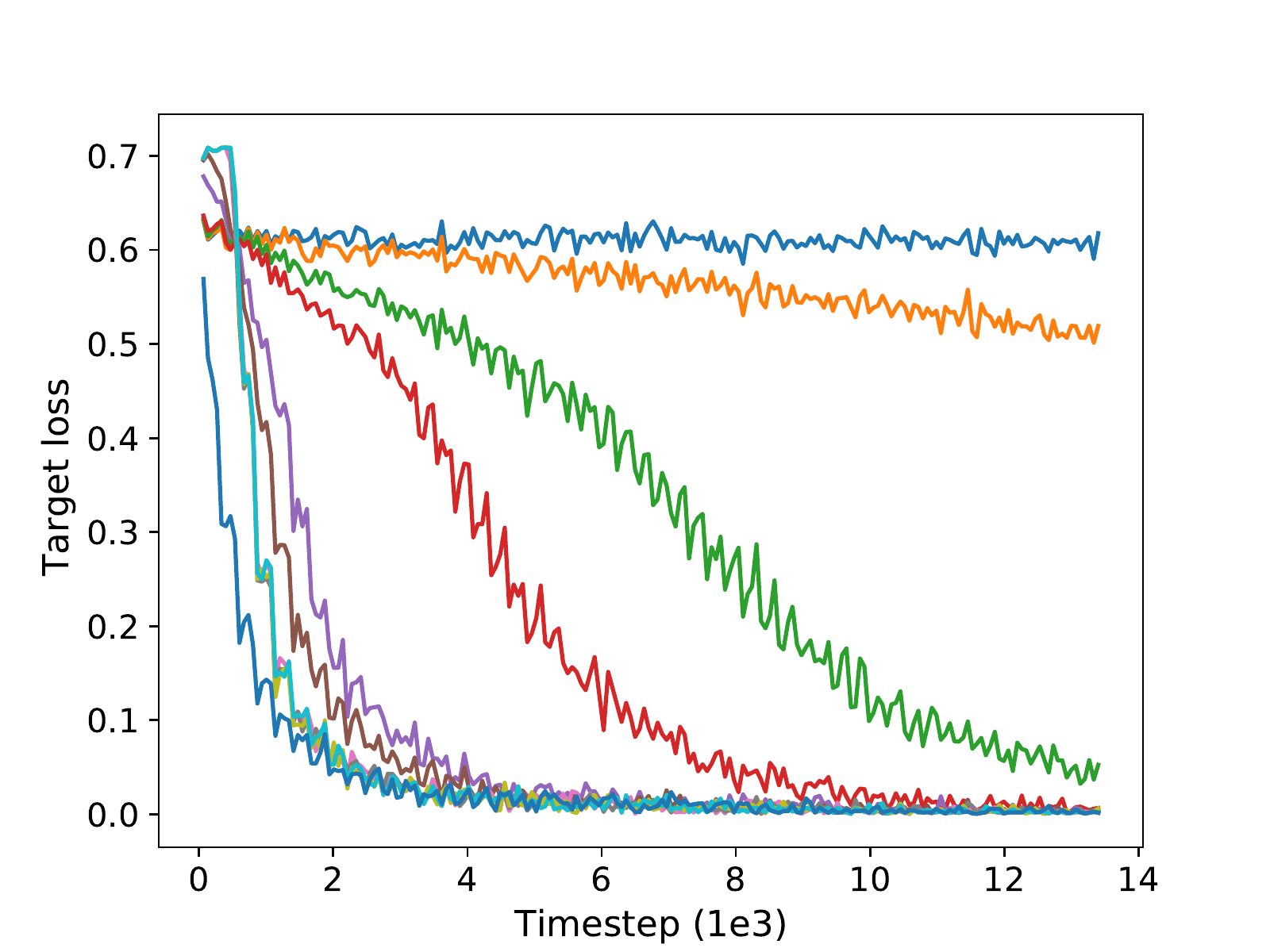}}\quad
	{\label{sfig:l}\includegraphics[scale=0.4]{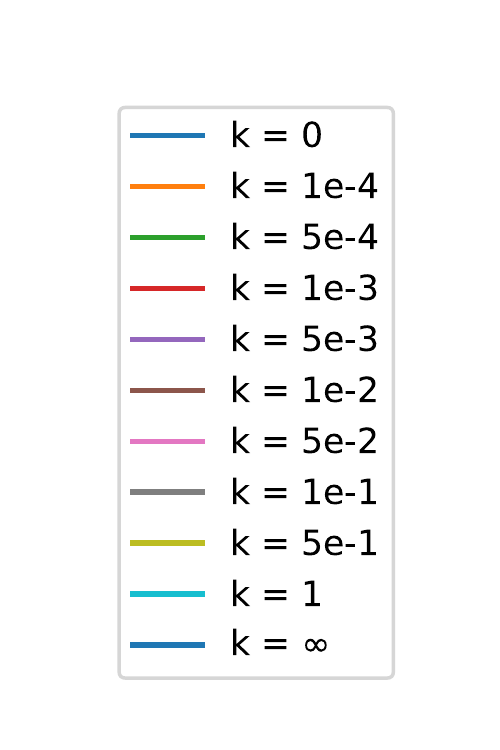}}\quad
	\subfloat[Knowledge forgetting from the source tasks] 
	{\label{sfig:b}\includegraphics[scale=0.4]{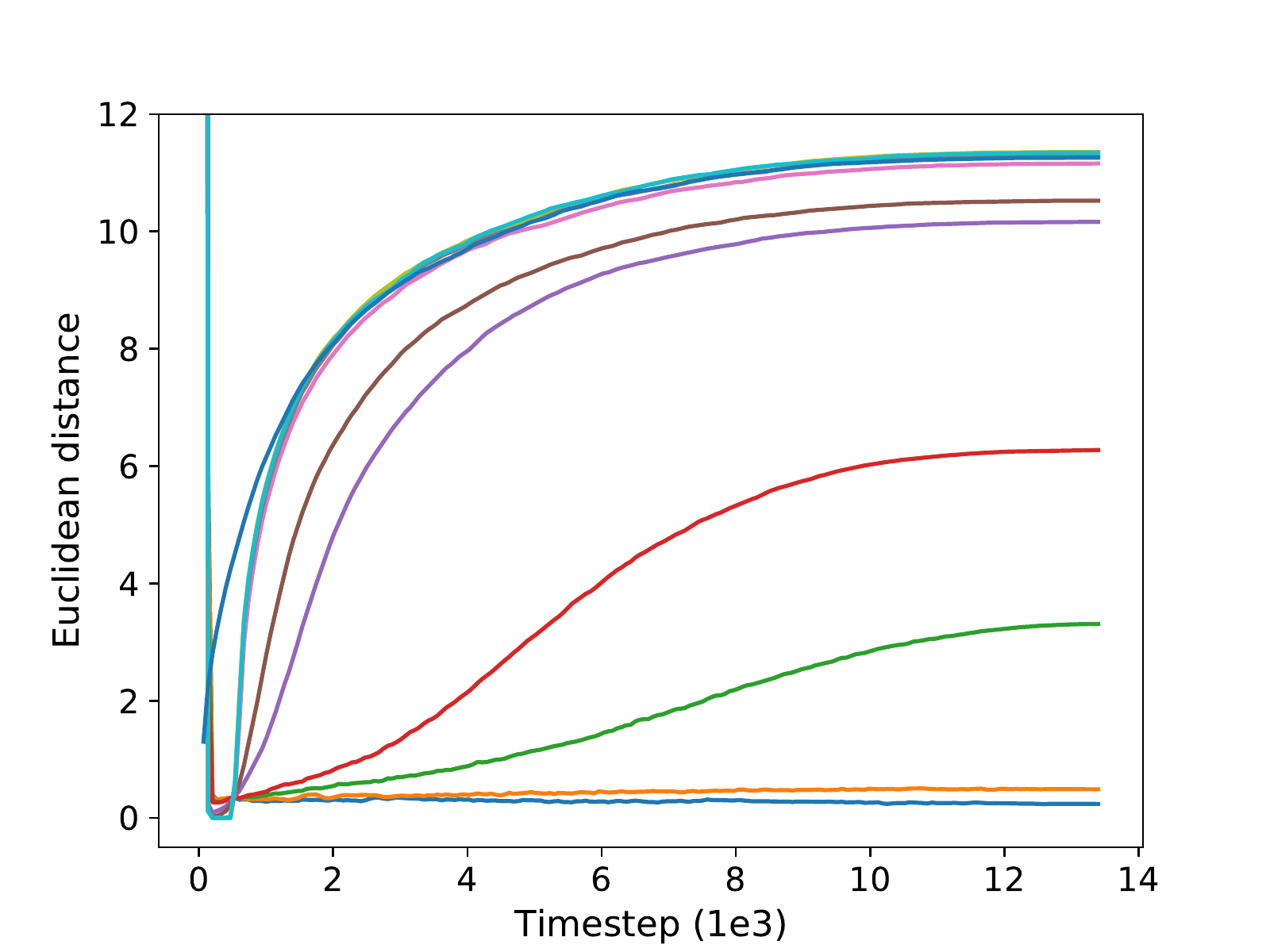}}\\	
	\caption{Learning curves obtained by BERT-base model trained with different objective shifting rate $k$ on CoLA task. We measure the knowledge forgetting by the Euclidean distance between the weights of the fine-tuning model and the pretrained model. With smaller $k$, the model achieves less knowledge forgetting from the source tasks while it takes more timesteps to converge on the target task.}\label{fig:forget_anay}
\end{figure*}

\paragraph{Forgetting Analysis}
As introduced in \S~\ref{ssec:obj_shift}, we realize multi-task fine-tuning with the Objective Shifting technique, which allows the model's learning objective to shift from the source tasks to the target tasks gradually. The hyperparameter $k$ controls the rate of the objective shifting.

Figure~\ref{fig:forget_anay} shows the learning curves of our fine-tuning methods with different $k$ value obtained by BERT-base model trained on CoLA dataset. As discussed in \S~\ref{ssec:obj_shift}, Fine-tuning and multi-task learning can be regarded as the special cases ($k \rightarrow \infty$ and $k \rightarrow 0$) of our method.

As shown in Figure~\ref{sfig:a}, with the larger shifting rate $k$, the model can converge quickly on the target task. As $k$ decreases, it takes a longer time for the model to converge on the target task because of the slower shifting from the pretrained knowledge recalling to target task learning.

Figure~\ref{sfig:b} shows the pretrained knowledge forgetting during the fine-tuning process. We measure the pretrained knowledge forgetting by the Euclidean distance between the weights of the fine-tuning model and the pretrained model. At the very early timesteps, the Euclidean distance drops sharply because of the random initialization and pretrained knowledge recalling. Then the curve rises with the growth rate slowing down because of the target task learning. As the objective shifting rate $k$ decreases, we find that the model can achieve less forgetting from the pretrained model at the end of the fine-tuning.

Overall, our methods provide a bridge between fine-tuning and multi-task learning. 
With smaller $k$, the model achieves less knowledge forgetting from the source tasks but risks not converging completely on the target task.
With a good balance between the pretrained knowledge recalling and new task learning, our methods can consistently outperform the vanilla fine-tuning by not only converging on target tasks but also less forgetting from source tasks.

\section{Related Works}
Catastrophic forgetting has been observed as a great challenge issue in sequential transfer learning, especially in the continuous learning paradigm \cite{mccloskey1989catastrophic,forget, forget2,cltutorial}. Many methods have been proposed to avoid catastrophic forgetting. 
Replay-based methods alleviate forgetting by relaying the samples of the previous tasks while learning the new task \cite{iCaRL,dgr,gem, gss}.
Parameter isolation-based methods avoid forgetting by updating a set of parameters for each task and freezing them for the new task \cite{packnet,hat,pnn,rcl,dan}.
Regularization-based methods propose to recall the previous knowledge with an extra regularization term \cite{kirkpatrick2017overcoming,si,imm,l2,mas,r-ewc,lwf,lfl,ebll,dmc}.
We focus on regularization-based methods in this paper, because they don't require the storage of the pretraining data, and more flexible compared to the parameter isolation-based methods.

Regularization-based methods can be further divided into data-focused and prior-focused methods. Data-focused methods regularize the new task learning by knowledge distillation from the pretrained model \cite{kd,lwf,lfl,ebll,dmc}. Prior-focused methods regard the distribution of the pretrained parameters as prior when learning the new task \cite{kirkpatrick2017overcoming,si,imm,l2,mas,r-ewc}. We adopted the idea of prior-focused methods because they enable the model to learn more general knowledge from the pretrained model's parameters more efficiently.
While the prior-focused methods, such as EWC \cite{kirkpatrick2017overcoming} and its variants \cite{ewc-v1,r-walk,r-ewc}, don't directly access to the pretraining data, they need some pretraining knowledge (e.g., Fisher information matrix of the source tasks), which is not available in our setting. 
Therefore, we further approximate to a quadratic penalty which is independent with the pretraining data given the pretrained parameters.

Catastrophic forgetting in NLP has raised increased attention recently \cite{howtraninnlp, emstudy, embarassmethod}. 
Many approaches have been proposed to overcome the forgetting problem in various domains, such as neural machine translation \cite{regtech,thompson2019overcoming} and reading comprehension \cite{xu2019forget}.
As sequential transfer learning widely used for NLP tasks \cite{howard2018universal,bert,roberta,albert,finetune-hou}, previous works explore many fine-tuning tricks to reduce catastrophic forgetting for adaptation of the deep pretrained LMs \cite{howard2018universal,sun2019fine,lee2019mixout,sidetune,chain-thaw}. In this paper, we bring the idea of multi-task learning which can inherently avoid catastrophic forgetting, apply it to the fine-tuning process with Pretraining Simulation and Objective Shifting mechanisms, and achieve consistent improvement with only the deep pretrained LMs available.

\section{Conclusion}
In this paper, we solve the catastrophic forgetting in transferring deep pretrained language models by bridging two transfer learning paradigm:
sequential fine-tuning and multi-task learning. 
To cope with the absence of pretraining data during the joint learning of pretraining task, 
we propose a Pretraining Simulation mechanism to learn the pretraining task without data. 
Then we propose the Objective Shifting mechanism to better balance the learning of the pretraining and downstream task. 
Experiments demonstrate the superiority of our method in the transferring of deep pretrained language models, and we provide the open-source \textsc{RecAdam} optimizer by integrating the proposed mechanisms into Adam optimizer to facilitate the better usage of deep pretrained language models.

\bibliography{recall_and_learn}

\begin{thebibliography}{69}
\expandafter\ifx\csname natexlab\endcsname\relax\def\natexlab#1{#1}\fi

\bibitem[{Aljundi et~al.(2018)Aljundi, Babiloni, Elhoseiny, Rohrbach, and
  Tuytelaars}]{mas}
Rahaf Aljundi, Francesca Babiloni, Mohamed Elhoseiny, Marcus Rohrbach, and
  Tinne Tuytelaars. 2018.
\newblock \href {https://doi.org/10.1007/978-3-030-01219-9\_9} {Memory aware
  synapses: Learning what (not) to forget}.
\newblock In \emph{Computer Vision - {ECCV} 2018 - 15th European Conference,
  Munich, Germany, September 8-14, 2018, Proceedings, Part {III}}, volume 11207
  of \emph{Lecture Notes in Computer Science}, pages 144--161. Springer.

\bibitem[{Aljundi et~al.(2019)Aljundi, Lin, Goujaud, and Bengio}]{gss}
Rahaf Aljundi, Min Lin, Baptiste Goujaud, and Yoshua Bengio. 2019.
\newblock \href {http://arxiv.org/abs/1903.08671} {Online continual learning
  with no task boundaries}.
\newblock \emph{CoRR}, abs/1903.08671.

\bibitem[{Arora et~al.(2019)Arora, Rahimi, and Baldwin}]{emstudy}
Gaurav Arora, Afshin Rahimi, and Timothy Baldwin. 2019.
\newblock \href {https://aclweb.org/anthology/papers/U/U19/U19-1011/} {Does an
  {LSTM} forget more than a cnn? an empirical study of catastrophic forgetting
  in {NLP}}.
\newblock In \emph{Proceedings of the The 17th Annual Workshop of the
  Australasian Language Technology Association, {ALTA} 2019, Sydney, Australia,
  December 4-6, 2019}, pages 77--86. Australasian Language Technology
  Association.

\bibitem[{Barone et~al.(2017)Barone, Haddow, Germann, and Sennrich}]{regtech}
Antonio Valerio~Miceli Barone, Barry Haddow, Ulrich Germann, and Rico Sennrich.
  2017.
\newblock \href {https://doi.org/10.18653/v1/d17-1156} {Regularization
  techniques for fine-tuning in neural machine translation}.
\newblock In \emph{Proceedings of the 2017 Conference on Empirical Methods in
  Natural Language Processing, {EMNLP} 2017, Copenhagen, Denmark, September
  9-11, 2017}, pages 1489--1494. Association for Computational Linguistics.

\bibitem[{Bentivogli et~al.(2009)Bentivogli, Magnini, Dagan, Dang, and
  Giampiccolo}]{rte}
Luisa Bentivogli, Bernardo Magnini, Ido Dagan, Hoa~Trang Dang, and Danilo
  Giampiccolo. 2009.
\newblock \href
  {https://tac.nist.gov/publications/2009/additional.papers/RTE5\_overview.proceedings.pdf}
  {The fifth {PASCAL} recognizing textual entailment challenge}.
\newblock In \emph{Proceedings of the Second Text Analysis Conference, {TAC}
  2009, Gaithersburg, Maryland, USA, November 16-17, 2009}. {NIST}.

\bibitem[{Bowman et~al.(2016)Bowman, Vilnis, Vinyals, Dai, J{\'{o}}zefowicz,
  and Bengio}]{svae}
Samuel~R. Bowman, Luke Vilnis, Oriol Vinyals, Andrew~M. Dai, Rafal
  J{\'{o}}zefowicz, and Samy Bengio. 2016.
\newblock \href {https://doi.org/10.18653/v1/k16-1002} {Generating sentences
  from a continuous space}.
\newblock In \emph{Proceedings of the 20th {SIGNLL} Conference on Computational
  Natural Language Learning, CoNLL 2016, Berlin, Germany, August 11-12, 2016},
  pages 10--21. {ACL}.

\bibitem[{Caruana(1997)}]{mtl1997}
Rich Caruana. 1997.
\newblock \href {https://doi.org/10.1023/A:1007379606734} {Multitask learning}.
\newblock \emph{Mach. Learn.}, 28(1):41--75.

\bibitem[{Cer et~al.(2017)Cer, Diab, Agirre, Lopez{-}Gazpio, and Specia}]{sts}
Daniel~M. Cer, Mona~T. Diab, Eneko Agirre, I{\~{n}}igo Lopez{-}Gazpio, and
  Lucia Specia. 2017.
\newblock \href {https://doi.org/10.18653/v1/S17-2001} {Semeval-2017 task 1:
  Semantic textual similarity multilingual and crosslingual focused
  evaluation}.
\newblock In \emph{Proceedings of the 11th International Workshop on Semantic
  Evaluation, SemEval@ACL 2017, Vancouver, Canada, August 3-4, 2017}, pages
  1--14. Association for Computational Linguistics.

\bibitem[{Chaudhry et~al.(2018)Chaudhry, Dokania, Ajanthan, and Torr}]{r-walk}
Arslan Chaudhry, Puneet~Kumar Dokania, Thalaiyasingam Ajanthan, and Philip
  H.~S. Torr. 2018.
\newblock \href {https://doi.org/10.1007/978-3-030-01252-6\_33} {Riemannian
  walk for incremental learning: Understanding forgetting and intransigence}.
\newblock In \emph{Computer Vision - {ECCV} 2018 - 15th European Conference,
  Munich, Germany, September 8-14, 2018, Proceedings, Part {XI}}, volume 11215
  of \emph{Lecture Notes in Computer Science}, pages 556--572. Springer.

\bibitem[{Chronopoulou et~al.(2019)Chronopoulou, Baziotis, and
  Potamianos}]{embarassmethod}
Alexandra Chronopoulou, Christos Baziotis, and Alexandros Potamianos. 2019.
\newblock \href {https://doi.org/10.18653/v1/n19-1213} {An embarrassingly
  simple approach for transfer learning from pretrained language models}.
\newblock In \emph{Proceedings of the 2019 Conference of the North American
  Chapter of the Association for Computational Linguistics: Human Language
  Technologies, {NAACL-HLT} 2019, Minneapolis, MN, USA, June 2-7, 2019, Volume
  1 (Long and Short Papers)}, pages 2089--2095. Association for Computational
  Linguistics.

\bibitem[{Dagan et~al.(2005)Dagan, Glickman, and Magnini}]{rte-1}
Ido Dagan, Oren Glickman, and Bernardo Magnini. 2005.
\newblock The pascal recognising textual entailment challenge.
\newblock In \emph{Machine Learning Challenges Workshop}, pages 177--190.
  Springer.

\bibitem[{Devlin et~al.(2019)Devlin, Chang, Lee, and Toutanova}]{bert}
Jacob Devlin, Ming{-}Wei Chang, Kenton Lee, and Kristina Toutanova. 2019.
\newblock \href {https://doi.org/10.18653/v1/n19-1423} {{BERT:} pre-training of
  deep bidirectional transformers for language understanding}.
\newblock In \emph{Proceedings of the 2019 Conference of the North American
  Chapter of the Association for Computational Linguistics: Human Language
  Technologies, {NAACL-HLT} 2019, Minneapolis, MN, USA, June 2-7, 2019, Volume
  1 (Long and Short Papers)}, pages 4171--4186. Association for Computational
  Linguistics.

\bibitem[{Dolan and Brockett(2005)}]{mrpc}
William~B. Dolan and Chris Brockett. 2005.
\newblock \href {https://www.aclweb.org/anthology/I05-5002/} {Automatically
  constructing a corpus of sentential paraphrases}.
\newblock In \emph{Proceedings of the Third International Workshop on
  Paraphrasing, IWP@IJCNLP 2005, Jeju Island, Korea, October 2005, 2005}. Asian
  Federation of Natural Language Processing.

\bibitem[{Felbo et~al.(2017)Felbo, Mislove, S{\o}gaard, Rahwan, and
  Lehmann}]{chain-thaw}
Bjarke Felbo, Alan Mislove, Anders S{\o}gaard, Iyad Rahwan, and Sune Lehmann.
  2017.
\newblock \href {https://doi.org/10.18653/v1/D17-1169} {Using millions of emoji
  occurrences to learn any-domain representations for detecting sentiment,
  emotion and sarcasm}.
\newblock In \emph{Proceedings of the 2017 Conference on Empirical Methods in
  Natural Language Processing}, pages 1615--1625, Copenhagen, Denmark.
  Association for Computational Linguistics.

\bibitem[{French(1999)}]{forget}
Robert~M. French. 1999.
\newblock \href {https://doi.org/https://doi.org/10.1016/S1364-6613(99)01294-2}
  {Catastrophic forgetting in connectionist networks}.
\newblock \emph{Trends in Cognitive Sciences}, 3(4):128 -- 135.

\bibitem[{Giampiccolo et~al.(2007)Giampiccolo, Magnini, Dagan, and
  Dolan}]{rte-3}
Danilo Giampiccolo, Bernardo Magnini, Ido Dagan, and Bill Dolan. 2007.
\newblock The third pascal recognizing textual entailment challenge.
\newblock In \emph{Proceedings of the ACL-PASCAL workshop on textual entailment
  and paraphrasing}, pages 1--9. Association for Computational Linguistics.

\bibitem[{Goodfellow et~al.(2013)Goodfellow, Mirza, Xiao, Courville, and
  Bengio}]{forget2}
Ian~J Goodfellow, Mehdi Mirza, Da~Xiao, Aaron Courville, and Yoshua Bengio.
  2013.
\newblock An empirical investigation of catastrophic forgetting in
  gradient-based neural networks.
\newblock \emph{arXiv preprint arXiv:1312.6211}.

\bibitem[{Hinton et~al.(2015)Hinton, Vinyals, and Dean}]{kd}
Geoffrey~E. Hinton, Oriol Vinyals, and Jeffrey Dean. 2015.
\newblock \href {http://arxiv.org/abs/1503.02531} {Distilling the knowledge in
  a neural network}.
\newblock \emph{CoRR}, abs/1503.02531.

\bibitem[{Hou et~al.(2019)Hou, Zhou, Liu, Wang, Che, Liu, and
  Liu}]{finetune-hou}
Yutai Hou, Zhihan Zhou, Yijia Liu, Ning Wang, Wanxiang Che, Han Liu, and Ting
  Liu. 2019.
\newblock Few-shot sequence labeling with label dependency transfer.
\newblock \emph{arXiv preprint arXiv:1906.08711}.

\bibitem[{Howard and Ruder(2018)}]{howard2018universal}
Jeremy Howard and Sebastian Ruder. 2018.
\newblock Universal language model fine-tuning for text classification.
\newblock \emph{arXiv preprint arXiv:1801.06146}.

\bibitem[{Husz{\'a}r(2017)}]{ewc-deriv}
Ferenc Husz{\'a}r. 2017.
\newblock On quadratic penalties in elastic weight consolidation.
\newblock \emph{arXiv preprint arXiv:1712.03847}.

\bibitem[{Jung et~al.(2016)Jung, Ju, Jung, and Kim}]{lfl}
Heechul Jung, Jeongwoo Ju, Minju Jung, and Junmo Kim. 2016.
\newblock \href {http://arxiv.org/abs/1607.00122} {Less-forgetting learning in
  deep neural networks}.
\newblock \emph{CoRR}, abs/1607.00122.

\bibitem[{Kingma and Ba(2015)}]{adam}
Diederik~P. Kingma and Jimmy Ba. 2015.
\newblock \href {http://arxiv.org/abs/1412.6980} {Adam: {A} method for
  stochastic optimization}.
\newblock In \emph{3rd International Conference on Learning Representations,
  {ICLR} 2015, San Diego, CA, USA, May 7-9, 2015, Conference Track
  Proceedings}.

\bibitem[{Kirkpatrick et~al.(2017)Kirkpatrick, Pascanu, Rabinowitz, Veness,
  Desjardins, Rusu, Milan, Quan, Ramalho, Grabska-Barwinska
  et~al.}]{kirkpatrick2017overcoming}
James Kirkpatrick, Razvan Pascanu, Neil Rabinowitz, Joel Veness, Guillaume
  Desjardins, Andrei~A Rusu, Kieran Milan, John Quan, Tiago Ramalho, Agnieszka
  Grabska-Barwinska, et~al. 2017.
\newblock Overcoming catastrophic forgetting in neural networks.
\newblock \emph{Proceedings of the national academy of sciences},
  114(13):3521--3526.

\bibitem[{Lan et~al.(2019)Lan, Chen, Goodman, Gimpel, Sharma, and
  Soricut}]{albert}
Zhenzhong Lan, Mingda Chen, Sebastian Goodman, Kevin Gimpel, Piyush Sharma, and
  Radu Soricut. 2019.
\newblock \href {http://arxiv.org/abs/1909.11942} {{ALBERT:} {A} lite {BERT}
  for self-supervised learning of language representations}.
\newblock \emph{CoRR}, abs/1909.11942.

\bibitem[{Lange et~al.(2019)Lange, Aljundi, Masana, Parisot, Jia, Leonardis,
  Slabaugh, and Tuytelaars}]{cltutorial}
Matthias~De Lange, Rahaf Aljundi, Marc Masana, Sarah Parisot, Xu~Jia, Ales
  Leonardis, Gregory~G. Slabaugh, and Tinne Tuytelaars. 2019.
\newblock \href {http://arxiv.org/abs/1909.08383} {Continual learning: {A}
  comparative study on how to defy forgetting in classification tasks}.
\newblock \emph{CoRR}, abs/1909.08383.

\bibitem[{Lee et~al.(2019)Lee, Cho, and Kang}]{lee2019mixout}
Cheolhyoung Lee, Kyunghyun Cho, and Wanmo Kang. 2019.
\newblock Mixout: Effective regularization to finetune large-scale pretrained
  language models.
\newblock \emph{arXiv preprint arXiv:1909.11299}.

\bibitem[{Lee et~al.(2017)Lee, Kim, Jun, Ha, and Zhang}]{imm}
Sang{-}Woo Lee, Jin{-}Hwa Kim, Jaehyun Jun, Jung{-}Woo Ha, and Byoung{-}Tak
  Zhang. 2017.
\newblock \href
  {http://papers.nips.cc/paper/7051-overcoming-catastrophic-forgetting-by-incremental-moment-matching}
  {Overcoming catastrophic forgetting by incremental moment matching}.
\newblock In \emph{Advances in Neural Information Processing Systems 30: Annual
  Conference on Neural Information Processing Systems 2017, 4-9 December 2017,
  Long Beach, CA, {USA}}, pages 4652--4662.

\bibitem[{Levesque et~al.(2012)Levesque, Davis, and Morgenstern}]{wnli}
Hector~J. Levesque, Ernest Davis, and Leora Morgenstern. 2012.
\newblock \href {http://www.aaai.org/ocs/index.php/KR/KR12/paper/view/4492}
  {The winograd schema challenge}.
\newblock In \emph{Principles of Knowledge Representation and Reasoning:
  Proceedings of the Thirteenth International Conference, {KR} 2012, Rome,
  Italy, June 10-14, 2012}. {AAAI} Press.

\bibitem[{Li et~al.(2018)Li, Grandvalet, and Davoine}]{l2}
Xuhong Li, Yves Grandvalet, and Franck Davoine. 2018.
\newblock Explicit inductive bias for transfer learning with convolutional
  networks.
\newblock \emph{arXiv preprint arXiv:1802.01483}.

\bibitem[{Li and Hoiem(2018)}]{lwf}
Zhizhong Li and Derek Hoiem. 2018.
\newblock \href {https://doi.org/10.1109/TPAMI.2017.2773081} {Learning without
  forgetting}.
\newblock \emph{{IEEE} Trans. Pattern Anal. Mach. Intell.}, 40(12):2935--2947.

\bibitem[{Liu et~al.(2018)Liu, Masana, Herranz, van~de Weijer, L{\'{o}}pez, and
  Bagdanov}]{r-ewc}
Xialei Liu, Marc Masana, Luis Herranz, Joost van~de Weijer, Antonio~M.
  L{\'{o}}pez, and Andrew~D. Bagdanov. 2018.
\newblock \href {https://doi.org/10.1109/ICPR.2018.8545895} {Rotate your
  networks: Better weight consolidation and less catastrophic forgetting}.
\newblock In \emph{24th International Conference on Pattern Recognition, {ICPR}
  2018, Beijing, China, August 20-24, 2018}, pages 2262--2268. {IEEE} Computer
  Society.

\bibitem[{Liu et~al.(2019)Liu, Ott, Goyal, Du, Joshi, Chen, Levy, Lewis,
  Zettlemoyer, and Stoyanov}]{roberta}
Yinhan Liu, Myle Ott, Naman Goyal, Jingfei Du, Mandar Joshi, Danqi Chen, Omer
  Levy, Mike Lewis, Luke Zettlemoyer, and Veselin Stoyanov. 2019.
\newblock \href {http://arxiv.org/abs/1907.11692} {Roberta: {A} robustly
  optimized {BERT} pretraining approach}.
\newblock \emph{CoRR}, abs/1907.11692.

\bibitem[{Lopez{-}Paz and Ranzato(2017)}]{gem}
David Lopez{-}Paz and Marc'Aurelio Ranzato. 2017.
\newblock \href
  {http://papers.nips.cc/paper/7225-gradient-episodic-memory-for-continual-learning}
  {Gradient episodic memory for continual learning}.
\newblock In \emph{Advances in Neural Information Processing Systems 30: Annual
  Conference on Neural Information Processing Systems 2017, 4-9 December 2017,
  Long Beach, CA, {USA}}, pages 6467--6476.

\bibitem[{Loshchilov and Hutter(2019)}]{adamw}
Ilya Loshchilov and Frank Hutter. 2019.
\newblock \href {https://openreview.net/forum?id=Bkg6RiCqY7} {Decoupled weight
  decay regularization}.
\newblock In \emph{7th International Conference on Learning Representations,
  {ICLR} 2019, New Orleans, LA, USA, May 6-9, 2019}. OpenReview.net.

\bibitem[{MacKay(2003)}]{laplace}
David J.~C. MacKay. 2003.
\newblock \emph{Information theory, inference, and learning algorithms}.
\newblock Cambridge University Press.

\bibitem[{Mallya and Lazebnik(2018)}]{packnet}
Arun Mallya and Svetlana Lazebnik. 2018.
\newblock \href {https://doi.org/10.1109/CVPR.2018.00810} {Packnet: Adding
  multiple tasks to a single network by iterative pruning}.
\newblock In \emph{2018 {IEEE} Conference on Computer Vision and Pattern
  Recognition, {CVPR} 2018, Salt Lake City, UT, USA, June 18-22, 2018}, pages
  7765--7773. {IEEE} Computer Society.

\bibitem[{Martens(2014)}]{efisher}
James Martens. 2014.
\newblock \href {http://arxiv.org/abs/1412.1193} {New perspectives on the
  natural gradient method}.
\newblock \emph{CoRR}, abs/1412.1193.

\bibitem[{McCloskey and Cohen(1989)}]{mccloskey1989catastrophic}
Michael McCloskey and Neal~J Cohen. 1989.
\newblock Catastrophic interference in connectionist networks: The sequential
  learning problem.
\newblock In \emph{Psychology of learning and motivation}, volume~24, pages
  109--165. Elsevier.

\bibitem[{Mou et~al.(2016)Mou, Meng, Yan, Li, Xu, Zhang, and
  Jin}]{howtraninnlp}
Lili Mou, Zhao Meng, Rui Yan, Ge~Li, Yan Xu, Lu~Zhang, and Zhi Jin. 2016.
\newblock \href {https://doi.org/10.18653/v1/d16-1046} {How transferable are
  neural networks in {NLP} applications?}
\newblock In \emph{Proceedings of the 2016 Conference on Empirical Methods in
  Natural Language Processing, {EMNLP} 2016, Austin, Texas, USA, November 1-4,
  2016}, pages 479--489. The Association for Computational Linguistics.

\bibitem[{Peters et~al.(2018)Peters, Neumann, Iyyer, Gardner, Clark, Lee, and
  Zettlemoyer}]{peters2018deep}
Matthew~E Peters, Mark Neumann, Mohit Iyyer, Matt Gardner, Christopher Clark,
  Kenton Lee, and Luke Zettlemoyer. 2018.
\newblock Deep contextualized word representations.
\newblock \emph{arXiv preprint arXiv:1802.05365}.

\bibitem[{Peters et~al.(2019)Peters, Ruder, and Smith}]{tuneornot}
Matthew~E. Peters, Sebastian Ruder, and Noah~A. Smith. 2019.
\newblock \href {https://doi.org/10.18653/v1/w19-4302} {To tune or not to tune?
  adapting pretrained representations to diverse tasks}.
\newblock In \emph{Proceedings of the 4th Workshop on Representation Learning
  for NLP, RepL4NLP@ACL 2019, Florence, Italy, August 2, 2019}, pages 7--14.
  Association for Computational Linguistics.

\bibitem[{Phang et~al.(2018)Phang, F{\'e}vry, and Bowman}]{smalldata}
Jason Phang, Thibault F{\'e}vry, and Samuel~R Bowman. 2018.
\newblock Sentence encoders on stilts: Supplementary training on intermediate
  labeled-data tasks.
\newblock \emph{arXiv preprint arXiv:1811.01088}.

\bibitem[{Rajpurkar et~al.(2016)Rajpurkar, Zhang, Lopyrev, and Liang}]{qnli}
Pranav Rajpurkar, Jian Zhang, Konstantin Lopyrev, and Percy Liang. 2016.
\newblock \href {https://doi.org/10.18653/v1/d16-1264} {Squad: 100, 000+
  questions for machine comprehension of text}.
\newblock In \emph{Proceedings of the 2016 Conference on Empirical Methods in
  Natural Language Processing, {EMNLP} 2016, Austin, Texas, USA, November 1-4,
  2016}, pages 2383--2392. The Association for Computational Linguistics.

\bibitem[{Rebuffi et~al.(2017)Rebuffi, Kolesnikov, Sperl, and Lampert}]{iCaRL}
Sylvestre{-}Alvise Rebuffi, Alexander Kolesnikov, Georg Sperl, and Christoph~H.
  Lampert. 2017.
\newblock \href {https://doi.org/10.1109/CVPR.2017.587} {icarl: Incremental
  classifier and representation learning}.
\newblock In \emph{2017 {IEEE} Conference on Computer Vision and Pattern
  Recognition, {CVPR} 2017, Honolulu, HI, USA, July 21-26, 2017}, pages
  5533--5542. {IEEE} Computer Society.

\bibitem[{Ribeiro et~al.(2019)Ribeiro, Melo, and Dias}]{ribeiro2019multi}
Joao Ribeiro, Francisco~S Melo, and Joao Dias. 2019.
\newblock Multi-task learning and catastrophic forgetting in continual
  reinforcement learning.
\newblock \emph{arXiv preprint arXiv:1909.10008}.

\bibitem[{Rosenfeld and Tsotsos(2020)}]{dan}
Amir Rosenfeld and John~K. Tsotsos. 2020.
\newblock \href {https://doi.org/10.1109/TPAMI.2018.2884462} {Incremental
  learning through deep adaptation}.
\newblock \emph{{IEEE} Trans. Pattern Anal. Mach. Intell.}, 42(3):651--663.

\bibitem[{Roy Bar-Haim and Szpektor.(2006)}]{rte-2}
Bill Dolan Lisa Ferro Danilo Giampiccolo Bernardo~Magnini Roy Bar-Haim,
  Ido~Dagan and Idan Szpektor. 2006.
\newblock The second pascal recognising textual entailment challenge.
\newblock In \emph{Proceedings of the second PASCAL challenges workshop on
  recognising textual entailment}, page 6–4.

\bibitem[{Ruder(2017)}]{mtl2017}
Sebastian Ruder. 2017.
\newblock \href {http://arxiv.org/abs/1706.05098} {An overview of multi-task
  learning in deep neural networks}.
\newblock \emph{CoRR}, abs/1706.05098.

\bibitem[{Ruder(2019)}]{ruder2019neural}
Sebastian Ruder. 2019.
\newblock \emph{Neural transfer learning for natural language processing}.
\newblock Ph.D. thesis, NUI Galway.

\bibitem[{Rusu et~al.(2016)Rusu, Rabinowitz, Desjardins, Soyer, Kirkpatrick,
  Kavukcuoglu, Pascanu, and Hadsell}]{pnn}
Andrei~A. Rusu, Neil~C. Rabinowitz, Guillaume Desjardins, Hubert Soyer, James
  Kirkpatrick, Koray Kavukcuoglu, Razvan Pascanu, and Raia Hadsell. 2016.
\newblock \href {http://arxiv.org/abs/1606.04671} {Progressive neural
  networks}.
\newblock \emph{CoRR}, abs/1606.04671.

\bibitem[{Schwarz et~al.(2018)Schwarz, Czarnecki, Luketina,
  Grabska{-}Barwinska, Teh, Pascanu, and Hadsell}]{ewc-v1}
Jonathan Schwarz, Wojciech Czarnecki, Jelena Luketina, Agnieszka
  Grabska{-}Barwinska, Yee~Whye Teh, Razvan Pascanu, and Raia Hadsell. 2018.
\newblock \href {http://proceedings.mlr.press/v80/schwarz18a.html} {Progress
  {\&} compress: {A} scalable framework for continual learning}.
\newblock In \emph{Proceedings of the 35th International Conference on Machine
  Learning, {ICML} 2018, Stockholmsm{\"{a}}ssan, Stockholm, Sweden, July 10-15,
  2018}, volume~80 of \emph{Proceedings of Machine Learning Research}, pages
  4535--4544. {PMLR}.

\bibitem[{Serr{\`{a}} et~al.(2018)Serr{\`{a}}, Suris, Miron, and
  Karatzoglou}]{hat}
Joan Serr{\`{a}}, Didac Suris, Marius Miron, and Alexandros Karatzoglou. 2018.
\newblock \href {http://proceedings.mlr.press/v80/serra18a.html} {Overcoming
  catastrophic forgetting with hard attention to the task}.
\newblock In \emph{Proceedings of the 35th International Conference on Machine
  Learning, {ICML} 2018, Stockholmsm{\"{a}}ssan, Stockholm, Sweden, July 10-15,
  2018}, volume~80 of \emph{Proceedings of Machine Learning Research}, pages
  4555--4564. {PMLR}.

\bibitem[{Shankar~Iyer and Csernai.(January 2017)}]{qqp}
Nikhil~Dandekar Shankar~Iyer and Kornl Csernai. January 2017.
\newblock { First quora dataset release: Question pairs}.
\newblock
  \url{https://www.quora.com/q/quoradata/First-Quora-Dataset-Release-Question-Pairs}.

\bibitem[{Shin et~al.(2017)Shin, Lee, Kim, and Kim}]{dgr}
Hanul Shin, Jung~Kwon Lee, Jaehong Kim, and Jiwon Kim. 2017.
\newblock \href
  {http://papers.nips.cc/paper/6892-continual-learning-with-deep-generative-replay}
  {Continual learning with deep generative replay}.
\newblock In \emph{Advances in Neural Information Processing Systems 30: Annual
  Conference on Neural Information Processing Systems 2017, 4-9 December 2017,
  Long Beach, CA, {USA}}, pages 2990--2999.

\bibitem[{Socher et~al.(2013)Socher, Perelygin, Wu, Chuang, Manning, Ng, and
  Potts}]{sst}
Richard Socher, Alex Perelygin, Jean Wu, Jason Chuang, Christopher~D. Manning,
  Andrew~Y. Ng, and Christopher Potts. 2013.
\newblock \href {https://www.aclweb.org/anthology/D13-1170/} {Recursive deep
  models for semantic compositionality over a sentiment treebank}.
\newblock In \emph{Proceedings of the 2013 Conference on Empirical Methods in
  Natural Language Processing, {EMNLP} 2013, 18-21 October 2013, Grand Hyatt
  Seattle, Seattle, Washington, USA, {A} meeting of SIGDAT, a Special Interest
  Group of the {ACL}}, pages 1631--1642. {ACL}.

\bibitem[{Sun et~al.(2019)Sun, Qiu, Xu, and Huang}]{sun2019fine}
Chi Sun, Xipeng Qiu, Yige Xu, and Xuanjing Huang. 2019.
\newblock How to fine-tune bert for text classification?
\newblock In \emph{China National Conference on Chinese Computational
  Linguistics}, pages 194--206. Springer.

\bibitem[{Thompson et~al.(2019)Thompson, Gwinnup, Khayrallah, Duh, and
  Koehn}]{thompson2019overcoming}
Brian Thompson, Jeremy Gwinnup, Huda Khayrallah, Kevin Duh, and Philipp Koehn.
  2019.
\newblock Overcoming catastrophic forgetting during domain adaptation of neural
  machine translation.
\newblock In \emph{Proceedings of the 2019 Conference of the North American
  Chapter of the Association for Computational Linguistics: Human Language
  Technologies, Volume 1 (Long and Short Papers)}, pages 2062--2068.

\bibitem[{Triki et~al.(2017)Triki, Aljundi, Blaschko, and Tuytelaars}]{ebll}
Amal~Rannen Triki, Rahaf Aljundi, Matthew~B. Blaschko, and Tinne Tuytelaars.
  2017.
\newblock \href {https://doi.org/10.1109/ICCV.2017.148} {Encoder based lifelong
  learning}.
\newblock In \emph{{IEEE} International Conference on Computer Vision, {ICCV}
  2017, Venice, Italy, October 22-29, 2017}, pages 1329--1337. {IEEE} Computer
  Society.

\bibitem[{Wang et~al.(2019)Wang, Singh, Michael, Hill, Levy, and Bowman}]{glue}
Alex Wang, Amanpreet Singh, Julian Michael, Felix Hill, Omer Levy, and
  Samuel~R. Bowman. 2019.
\newblock \href {https://openreview.net/forum?id=rJ4km2R5t7} {{GLUE:} {A}
  multi-task benchmark and analysis platform for natural language
  understanding}.
\newblock In \emph{7th International Conference on Learning Representations,
  {ICLR} 2019, New Orleans, LA, USA, May 6-9, 2019}. OpenReview.net.

\bibitem[{Warstadt et~al.(2019)Warstadt, Singh, and Bowman}]{cola}
Alex Warstadt, Amanpreet Singh, and Samuel~R. Bowman. 2019.
\newblock \href {https://transacl.org/ojs/index.php/tacl/article/view/1710}
  {Neural network acceptability judgments}.
\newblock \emph{{TACL}}, 7:625--641.

\bibitem[{Williams et~al.(2018)Williams, Nangia, and Bowman}]{mnli}
Adina Williams, Nikita Nangia, and Samuel~R. Bowman. 2018.
\newblock \href {https://doi.org/10.18653/v1/n18-1101} {A broad-coverage
  challenge corpus for sentence understanding through inference}.
\newblock In \emph{Proceedings of the 2018 Conference of the North American
  Chapter of the Association for Computational Linguistics: Human Language
  Technologies, {NAACL-HLT} 2018, New Orleans, Louisiana, USA, June 1-6, 2018,
  Volume 1 (Long Papers)}, pages 1112--1122. Association for Computational
  Linguistics.

\bibitem[{Xu and Zhu(2018)}]{rcl}
Ju~Xu and Zhanxing Zhu. 2018.
\newblock \href
  {http://papers.nips.cc/paper/7369-reinforced-continual-learning} {Reinforced
  continual learning}.
\newblock In \emph{Advances in Neural Information Processing Systems 31: Annual
  Conference on Neural Information Processing Systems 2018, NeurIPS 2018, 3-8
  December 2018, Montr{\'{e}}al, Canada}, pages 907--916.

\bibitem[{Xu et~al.(2019)Xu, Zhong, Yepes, and Lau}]{xu2019forget}
Ying Xu, Xu~Zhong, Antonio Jose~Jimeno Yepes, and Jey~Han Lau. 2019.
\newblock Forget me not: Reducing catastrophic forgetting for domain adaptation
  in reading comprehension.
\newblock \emph{arXiv preprint arXiv:1911.00202}.

\bibitem[{Xue et~al.(2019)Xue, Han, Zheng, Gao, and Guo}]{xue2019multi}
Jiabin Xue, Jiqing Han, Tieran Zheng, Xiang Gao, and Jiaxing Guo. 2019.
\newblock A multi-task learning framework for overcoming the catastrophic
  forgetting in automatic speech recognition.
\newblock \emph{arXiv preprint arXiv:1904.08039}.

\bibitem[{Yang et~al.(2019)Yang, Dai, Yang, Carbonell, Salakhutdinov, and
  Le}]{xlnet}
Zhilin Yang, Zihang Dai, Yiming Yang, Jaime~G. Carbonell, Ruslan Salakhutdinov,
  and Quoc~V. Le. 2019.
\newblock \href
  {http://papers.nips.cc/paper/8812-xlnet-generalized-autoregressive-pretraining-for-language-understanding}
  {Xlnet: Generalized autoregressive pretraining for language understanding}.
\newblock In \emph{Advances in Neural Information Processing Systems 32: Annual
  Conference on Neural Information Processing Systems 2019, NeurIPS 2019, 8-14
  December 2019, Vancouver, BC, Canada}, pages 5754--5764.

\bibitem[{Zenke et~al.(2017)Zenke, Poole, and Ganguli}]{si}
Friedemann Zenke, Ben Poole, and Surya Ganguli. 2017.
\newblock \href {http://proceedings.mlr.press/v70/zenke17a.html} {Continual
  learning through synaptic intelligence}.
\newblock In \emph{Proceedings of the 34th International Conference on Machine
  Learning, {ICML} 2017, Sydney, NSW, Australia, 6-11 August 2017}, volume~70
  of \emph{Proceedings of Machine Learning Research}, pages 3987--3995. {PMLR}.

\bibitem[{Zhang et~al.(2019{\natexlab{a}})Zhang, Sax, Zamir, Guibas, and
  Malik}]{sidetune}
Jeffrey~O. Zhang, Alexander Sax, Amir~Roshan Zamir, Leonidas~J. Guibas, and
  Jitendra Malik. 2019{\natexlab{a}}.
\newblock \href {http://arxiv.org/abs/1912.13503} {Side-tuning: Network
  adaptation via additive side networks}.
\newblock \emph{CoRR}, abs/1912.13503.

\bibitem[{Zhang et~al.(2019{\natexlab{b}})Zhang, Zhang, Ghosh, Li, Tasci, Heck,
  Zhang, and Kuo}]{dmc}
Junting Zhang, Jie Zhang, Shalini Ghosh, Dawei Li, Serafettin Tasci, Larry~P.
  Heck, Heming Zhang, and C.{-}C.~Jay Kuo. 2019{\natexlab{b}}.
\newblock \href {http://arxiv.org/abs/1903.07864} {Class-incremental learning
  via deep model consolidation}.
\newblock \emph{CoRR}, abs/1903.07864.

\end{thebibliography}
\bibliographystyle{acl_natbib}

\end{document}